\documentclass[12pt]{article}

\usepackage{times}
\usepackage{epsfig}
\usepackage{graphicx}
\usepackage{amsmath}
\usepackage{amssymb}
\usepackage{csquotes}
\usepackage{amssymb}
\usepackage{tcolorbox}
\usepackage{xcolor}
\usepackage{listings}
\usepackage{pgfplots}
\usepackage{listings}
\usepackage{xcolor}
\usepackage{pgf-pie}
\usepackage{amsmath, amssymb}
\usepackage{graphicx}
\tcbuselibrary{breakable}
\usepackage{setspace}
\usepackage{pgfplots}
\pgfplotsset{compat=1.18}
\usepackage{xcolor}
\usepackage{caption}
\usepackage{tikz}
\usetikzlibrary{trees, positioning}
\usepackage{graphicx}
\usepackage{forest}
\usepackage{tikz-qtree}

\usepackage[english]{babel}

\usepackage{parskip}

\usepackage[breaklinks=true,letterpaper=true,colorlinks,bookmarks=false, allcolors=blue]{hyperref}

\usepackage[a4paper, margin={1in}]{geometry}

\usepackage[style=ieee, maxnames=3, backend=biber]{biblatex}
\usepackage{float} 

\begin{document}

\clearpage\thispagestyle{empty}

\title{IE4100 BENG Dissertation: Large Language Models for Supply Chain Resilience Management}

\author{
    {Guoyun Zhang\\
    {\tt\small zhang.guoyun@u.nus.edu}}\\
    \\
    \begin{center}
        {National University of Singapore,
         21 Lower Kent Ridge Rd, Singapore 119077}
    \end{center} \
}

\doublespacing
\begin{titlepage}
    \centering
    \vspace*{1cm}

    {\Large \bfseries IE4100 BENG Dissertation}

    \vspace{1.5cm}

    {\bfseries Large Language Model enabled Mathematical Modeling}

    \vspace{1.5cm}

    {Zhang Guoyun}

    \vspace{1.5cm}

    {\textit{DEPARTMENT OF INDUSTRIAL SYSTEMS ENGINEERING AND MANAGEMENT\\NATIONAL UNIVERSITY OF SINGAPORE}}

    \vspace{1.5cm}

    {\textit{AY 2024/2025 Semester 1\&2}}
   
\end{titlepage}
\clearpage
\newpage

\clearpage\thispagestyle{empty} 
\begin{centering}
\section*{Abstract}
\end{centering}
\thispagestyle{empty}
The integration of Large Language Models (LLMs) into supply chain optimization presents a promising frontier for enhancing decision-making capabilities in operations research (OR). Traditional optimization approaches—such as linear programming, mixed integer programming, and simulation—require domain expertise to convert real-world problems into solvable mathematical formulations. Despite the power of optimization solvers like Gurobi and COPT, human expertise remains indispensable in defining objectives, constraints, and variable structures. This research explores the capacity of LLMs, particularly the DeepSeek-R1 model, to bridge this formulation gap through natural language understanding and code generation.

While previous studies using models like GPT-4, Claude, and Bard demonstrated significant advancements in various NLP and reasoning tasks, high token costs and model hallucinations hinder practical deployment in supply chain contexts. In response, the economically efficient and high-performing DeepSeek-R1 model—equipped with reinforcement learning—has emerged as a competitive alternative. Despite its strong performance in benchmarks like LiveCodeBench and Math-500, limited research has explored its utility in real-world OR problem-solving, particularly under hallucination-prone conditions.

This study systematically evaluates DeepSeek-R1 across four critical OR benchmarks: NL4OPT, IndustryOR, EasyLP, and ComplexOR. Our methodology includes baseline evaluation, hallucination taxonomy development, and the implementation of mitigation strategies such as LLM-as-a-Judge, Few-shot Learning (FSL), Tool Calling, and a Multi-agent Framework. These strategies are designed to reduce code hallucination, improve problem formulation accuracy, and align model output with user intent.

The taxonomy categorizes hallucinations into three primary types: Attribute Errors, Syntax Errors, and Logical Errors. Attribute Errors, comprising over 65\% of observed hallucinations, often stem from incorrect or fabricated API calls. We embedded a RAG (Retrieval-Augmented Generation) tool using static API documentation in .pyi format to guide the model during tool referencing and reduce misuse. Nonetheless, hallucinations persisted when the model failed to consult the tool or misinterpreted its output.

Among all mitigation techniques, LLM-as-a-Judge achieved the most significant improvement across benchmarks, increasing NL4OPT accuracy from 78.8\% to 92.3\% and IndustryOR from 37.0\% to 50.0\%. It operates by having the model review its own output, critique the logic and code, and regenerate more accurate formulations. Few-shot Learning contributed marginal improvements but faced diminishing returns on complex datasets like ComplexOR due to insufficient generalization. The Multi-agent Framework, involving separate "Mathematician" and "Coder" agents powered by DeepSeek-R1, struggled with accuracy when upstream outputs were flawed. Lastly, Tool Calling, while conceptually effective in mitigating API misuse, faced challenges in tool invocation frequency and model overconfidence.

In conclusion, DeepSeek-R1, when enhanced with LLM-as-a-Judge, demonstrates substantial potential in supporting optimization tasks across diverse supply chain scenarios. The study not only offers a structured pipeline for using LLMs in OR but also highlights the challenges of hallucination, tool integration, and model interpretability. This work serves as a foundation for future research on aligning LLM outputs with domain-specific correctness, and for industrial applications seeking cost-effective, AI-augmented optimization solutions.
\thispagestyle{empty}
\clearpage

\clearpage\thispagestyle{empty}
\begin{centering}
\section*{Acknowledgement}
\end{centering}
I would like to express my deepest gratitude to my supervisor, Professor Qin Han Zhang, for his invaluable guidance, patience, and unwavering support throughout the research and writing of this thesis. His insightful advice and constructive feedback have been instrumental in shaping this work.

I would also like to extend my heartfelt appreciation to my supervisor Yang Shan Shan at A*STAR, whose industry knowledge and encouragement have greatly contributed to my academic growth.

A special thanks to my fellow coursemates and friends for their moral support, collaboration, and countless discussions that have enriched my understanding of the subject.

Finally, I am profoundly grateful to my partner and my family for their unconditional love, encouragement, and sacrifices that have enabled me to reach this huge milestone. Their belief in me has been my greatest motivation.

This thesis would not have been possible without the contributions and support of all these individuals. Thank you!
\thispagestyle{empty}
\clearpage
\newpage

\clearpage\thispagestyle{empty}
\tableofcontents  
\thispagestyle{empty}
\clearpage
\newpage  

\section{Introduction}\setcounter{page}{1}
In real-world scenarios, companies in the supply chain sector often face complex operational challenges due to factors such as geographical dispersion, product variety, and varying levels of customization \cite{bozarth2009impact}.

Traditionally, Operations Research (OR) methods—including linear programming, mixed-integer linear programming (MILP), queuing theory, and simulations—are employed to formulate and solve these problems. The Institute for Operations Research and the Management Sciences (INFORMS) defines OR as “the discipline of applying advanced analytical methods to help make better decisions” \cite{manson2006operations}. It further explains: “By using techniques such as mathematical modelling to analyze complex situations, operations research gives executives the power to make more effective decisions and build more productive systems.”

Solving OR problems typically involves three main steps: (1) identifying the problem and objectives, (2) formulating the problem mathematically, and (3) optimizing the model to find the best solution.

With the advancement of optimization software tools such as Gurobi and CPLEX, it is now possible to efficiently solve large-scale mathematical formulations. However, these tools cannot independently identify key variables or formulate problems from natural language descriptions. As a result, human expertise is still essential in defining model structures and translating real-world situations into mathematical terms. To address this limitation, Large Language Models (LLMs) have emerged as promising tools that can assist with problem formulation by interpreting natural language input and generating corresponding code or mathematical expressions.

Recent advances in LLMs offer the potential to bridge the gap between human reasoning and supply chain automation. These models can support users by converting natural language requirements into executable code. Most modern LLMs are based on decoder-only Transformer architectures and are trained on large-scale datasets using self-supervised learning to perform tasks like code generation, text extraction, and reasoning.

Popular closed-source LLMs such as ChatGPT-4o (OpenAI, 2024), Claude 3.7 (Anthropic, 2025), and Gemini 2.5 (Google, 2025) continue to evolve rapidly, backed by substantial computational and financial resources \cite{guo2025deepseek}.
\begin{figure}[h]
    \centering
    \includegraphics[width=0.5 \textwidth]{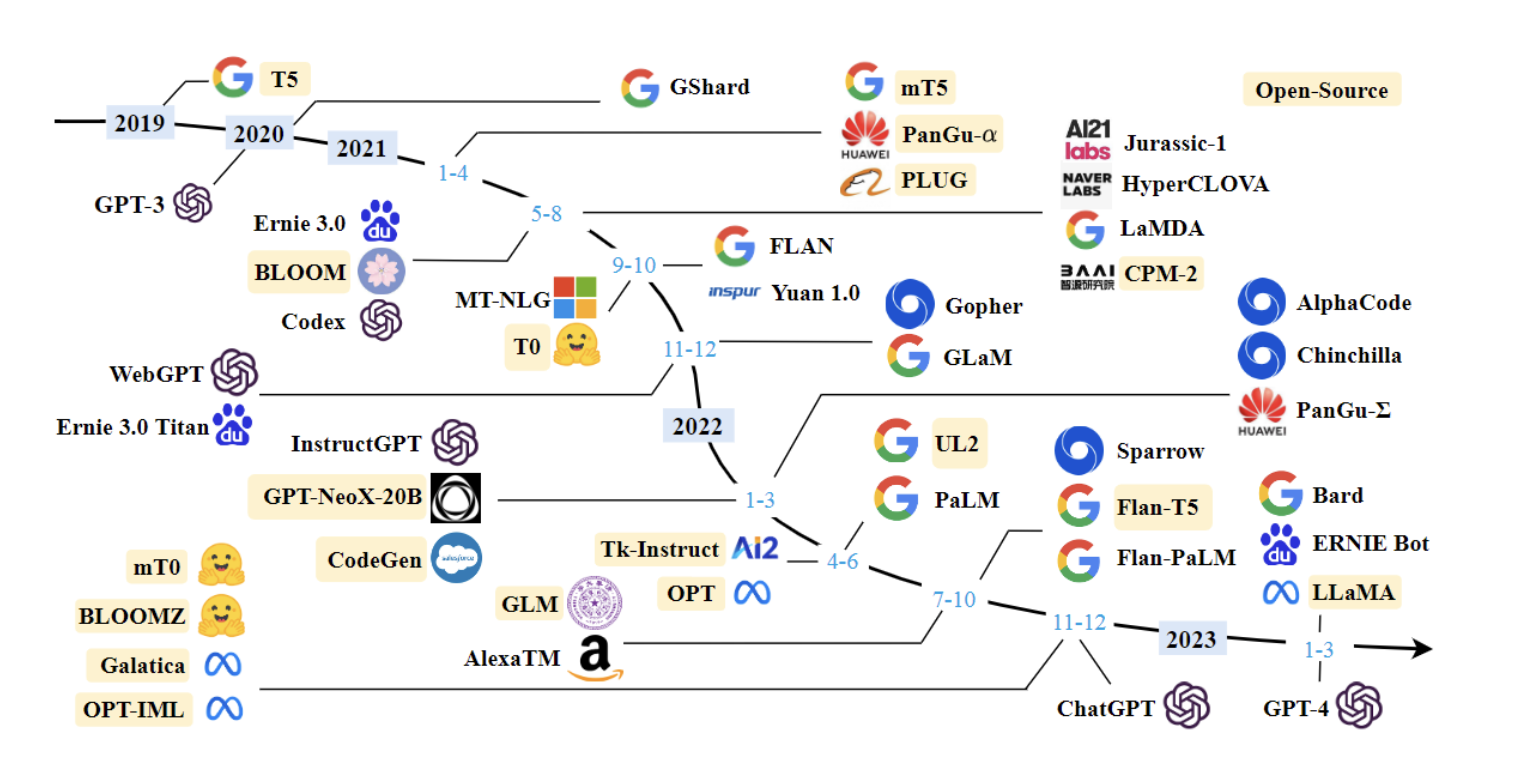}
    \caption{Evolution of LLMs \cite{zhao2023survey}}
\end{figure}
Currently, GPT-4 demonstrates strong performance on programming benchmarks such as HumanEval and MBPP. LLaMA-3-8B, for instance, achieves 67.7\% accuracy on HumanEval \cite{zhang2024llm}. In the domain of OR, GPT-4 has also achieved outstanding results on the NL4OPT and IndustryOR benchmarks. However, the cost of using state-of-the-art LLMs remains a barrier. GPT-4o, for example, ranges from \$5 to \$80 per 1 million prompt tokens, which makes extensive experimentation and ablation studies economically infeasible.

This changed with the introduction of the DeepSeek model, which offers state-of-the-art accuracy at a fraction of the cost. DeepSeek-R1 incorporates reinforcement learning techniques to encourage more thoughtful generation. It outperforms models like Claude and GPT-4o across multiple domains, including English, programming, mathematics, and Chinese. Notably, it performs particularly well on the LiveCodeBench (coding) and Math-500 (mathematics) benchmarks, both of which are relevant to this study. Moreover, DeepSeek-R1 is significantly more affordable—costing between \$0.035 and \$1.10 per million tokens—making it over 50 times cheaper than GPT-4. However, as the model is relatively new, limited literature exists on its evaluation, especially in OR applications.

Despite the remarkable progress of LLMs, they still face challenges when applied to real-world problem-solving. First, the inner workings of LLMs are still not fully understood, making them somewhat of a “black box.” Investigating the factors behind their strong performance remains an open research question. Second, since most models are trained by private companies using proprietary data, publicly available training data is limited. This can hinder the model's ability to generalize to specific industrial problems. Finally, hallucination—where the model produces incorrect or fabricated outputs—is a major concern. This can occur during both the formulation of mathematical models and the code generation process, potentially leading to unsolvable or inaccurate OR problems \cite{zhang2024llm}. While existing models demonstrate promising performance, addressing these limitations is crucial for improving their reliability and practical use.

The goal of this project is to identify the limitations of the DeepSeek-R1 model in code generation for complex OR problems. We evaluate the model using four benchmarks: IndustryOR, EasyLP, NL4OPT, and ComplexOR. Additionally, this research investigates how variations in problem formulation affect model performance and aims to provide users with templates to improve input quality and reduce inefficiencies in supply chain applications.

We also investigate the use of Tool Calling, a technique that builds upon Retrieval-Augmented Generation (RAG), to improve the accuracy and reliability of LLM outputs. RAG enhances model performance by retrieving relevant information from an external knowledge base before generating a response, allowing the model to incorporate domain-specific context without requiring additional training. Tool Calling extends this concept by enabling the model to actively interact with predefined functions or APIs during generation, further reducing hallucination and guiding it toward accurate solutions. This combined approach allows us to tailor general-purpose LLMs to specialized domains such as operations research (OR), using structured documentation and code references. Ultimately, this research aims to propose a more robust and cost-efficient LLM-based framework for solving OR problems, while also identifying the limitations and potential areas for improvement in models like DeepSeek-R1.

\subsection{Large Language Model (LLM)}

Recent progress in Natural Language Processing (NLP) has led to the development of LLMs. In just a few years, the LLMs have evolved incredibly, going from non-existent to ubiquitous in the machine learning field.

While growing very rapidly, the basic building blocks behind LLMs (GPT-2 \cite{radford2019language}, Gemma 2 \cite{gemmateam2024gemma2}, Llama 3 \cite{dubey2024llama3}, and etc.) have never changed. They are mainly composed of the following components:

\begin{figure}[h]
    \centering
    \includegraphics[width=0.5 \textwidth]{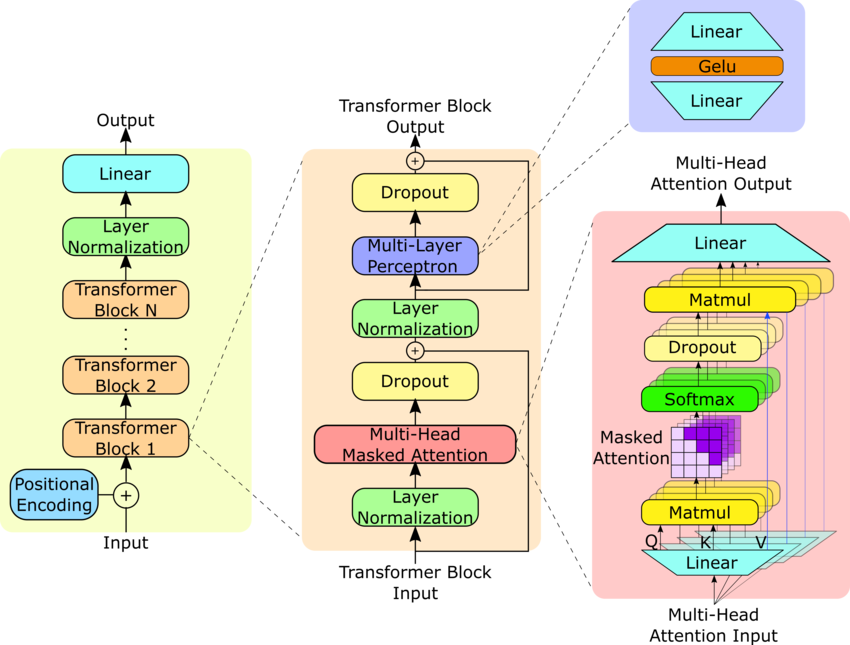}
    \caption{\textit{GPT-2} Architecture \cite{yang2023fluid}}
\end{figure}

\begin{itemize}
    \item \textbf{Embedding}: It transforms a list of tokenized words to a list of embeddings (vectors) that represent the words in a high dimensional vector space.
    \item \textbf{Positional Encoding}: In contrast to RNNs and LSTMs \cite{hochreiter1997lstm}, which implicitly capture token order through their recurrent structure, LLMs introduce a positional encoding layer after embedding to add positional information to the list of embeddings they will process. Commonly used positional encodings include absolute position encodings \cite{vaswani2023attention} and relative position embeddings \cite{huang2020improve}.
    \item \textbf{Attention}: An attention layer takes in a query matrix $Q$ and a key matrix $K$, and computes how each query vector $q \in Q$ attends to the vectors $k \in K$. Essentially, it calculates a similarity score, usually using the dot product, between each $q$ and $k$. These similarity scores are then used as coefficients to compute a linear combination of the corresponding value vectors from the value matrix $V$, resulting in the vector that $q$ attends to. Attention layers are usually used in a multi-head manner, meaning that the input embeddings are projected into several subspaces (heads), with each head applying its own attention mechanism. The outputs from all heads are then concatenated and combined. Recently, a new way to use attention, called GQA \cite{ainslie2023gqa}, is proposed, achieving comparable performance to MHA (multi-head attention) while being faster.
\end{itemize}

With the seemingly simple building blocks above, LLMs showed remarkable versatility, demonstrating their potential in a variety of fields, including medicine \cite{singhal2023large} \cite{agrawal2022large} \cite{you2020bertmesh}, law \cite{yu2022legal} \cite{peric2020legal} and computer programming \cite{chen2021evaluating} \cite{fried2023incoder}. At the same time, its ability to operate in different languages (natural languages, programming languages, and other domain-specific languages) allows it to connect with other software tools, thus greatly extending its performance in a variety of tasks, from writing code \cite{gao2023pal} to solving maths \cite{schick2023toolformer} \cite{heyueya2023solving} and optimization problems \cite{ahmaditeshnizi2023optimus}.

However, LLMs are not powerful without limitations. \cite{longchat2023how} pointed out that the performance of many open source long-context LLMs degrades severely as the input length increases. Moreover, \cite{liu2024lost} found that the performance of LLMs depends on the location of the relevant information: performance tends to be highest when the relevant information occurs at the beginning or end of the input context, and drops significantly when the relevant information is in the middle of a long context, even with explicitly long context models. Specifically for maths problems, \cite{xie2024mathlearner} concluded that LLMs tend to match problem-solving strategies based on textual similarities rather than truly understanding the principles of problems, and even CoT \cite{wei2023chainofthoughtpromptingelicitsreasoning} or specialized training cannot fully adjust the inherent incorrect problem solving methods of LLMs. In the start of 2025, Deepseek was introduced as a Reinforcement Learning LLM incorporated with Group Relative Policy Optimization method. It is unique from the other mainstream LLM such as GPT-4o, LLaMa and etc.





\subsection{Optimization in Supply Chain}
Supply chains are integrated networks of facilities and transportation options that manage the supply, manufacture, storage and distribution of materials and products. They vary considerably in size, complexity and scale from industry to industry \cite{simchi2005logic}.

In the impact of supply chain complexity on manufacturing plant performance \cite{bozarth2009impact}, there are three primary sources of supply chain complexity that can significantly affect manufacturing plant performance:

\begin{itemize}
    \item \textbf{Internal manufacturing complexity.} It is defined as the level of detail and dynamic complexity found within the manufacturing facility’s products, processes, and planning and control systems. Potential drivers of internal manufacturing complexity include the number of supported parts and products, the types of manufacturing processes, and the stability of manufacturing schedules from one period to the next \cite{flynn1999information}.
    \item \textbf{Downstream complexity.} Downstream complexity is the level of detail and dynamic complexity originating in a manufacturing facility’s downstream markets. The potential drivers of downstream complexity include the number of customers, the heterogeneity of customer needs, the average length of the product life cycle and the variability of demand.
    \item \textbf{Upstream complexity.} It is characterized by the level of detail and dynamic complexity originating in a manufacturing facility’s supply base. Potential drivers of upstream complexity include the number of supplier relationships that must be managed, the delivery lead time and reliability of suppliers, and the extent of global sourcing.
\end{itemize}

These complexities can be formulated as objective functions and optimized through operation research approaches. The following is a formal definition of the general form of the optimization problem:
\begin{singlespace}
\begin{tcolorbox}[
    title=\textbf{General Optimization Objective Function},
    colframe=black!50, colback=black!5,
    breakable  
]
\textbf{Objective Function:}
\[
\underset{x_1,...x_n}{\text{maximize}} \quad z = \textbf{c}^T \textbf{x}
\]

\textbf{Subject to:}
\[
\textbf{Ax} = \textbf{b}\]
\[
\textbf{x} \geq \textbf{0}
\]
\end{tcolorbox}
\end{singlespace}

We will explain the concept with a simple example from 
NL4OPT\cite{ramamonjison2022nl4opt}.
\begin{singlespace}
\begin{tcolorbox}[
    title=\textbf{A NL4OPT example},
    colframe=black!50, colback=black!5,
    breakable  
]
An office supply company makes two types of printers: color printers $X_1$ and black and white printers $X_2$. Different sections of the factory with different teams produce each printer. The color printer team can produce at most 20 color printers per day while the black and white printer team can produce at most 30 black and white printers per day. Both teams require use of the same paper tray installing machine and this machine can make at most 35 printers of either type each day. Color printers generate a profit of \$200 per printer while black and white printers generate a profit of \$70 per printer. How many of each printer should be made to maximize the company's profit?
Based on the definition of the optimization problem, the above problem can then be formulated as:
\textbf{Objective Function:}
\[
\underset{X_1, X_2}{\text{maximize}} \quad z = 200X_1 + 70X_2
\]

\textbf{Subject to:}
\[
\begin{cases}
    X_1 \leq 20 & \text{(cost constraint for } X_1 \text{)} \\
    X_2 \leq 30 & \text{(cost constraint for } X_2 \text{)} \\
    X_1 + X_2 \leq 35 & \text{(printer constraint)} \\
    X_1, X_2 \geq 0 & \text{(non-negativity constraint)}
\end{cases}
\]

\hfill
\end{tcolorbox}
\end{singlespace}

\subsubsection{IndustryOR}

IndustryOR is the first benchmark specifically designed to evaluate the performance of large language models on real-world industrial operations research (OR) problems. It consists of 100 authentic OR tasks and spans five common problem types: linear programming, integer programming, mixed-integer programming, nonlinear programming, and other specialized formulations. These problems are categorized into three levels of difficulty—easy, medium, and difficult—to assess model performance across a range of complexities.

Each entry in the benchmark includes: (1) a natural language prompt that describes the OR problem, (2) an execution result field that outlines the reasoning process generated by the LLM, and (3) an execution state, which indicates whether the model successfully solved the problem. This benchmark provides a comprehensive framework for evaluating how well LLMs can interpret, formulate, and solve industrial-scale optimization tasks\cite{huang2024orlm}.

\subsubsection{NL4OPT}
In addition to IndustryOR, we also utilized the Natural Language for Optimization Competition (NL4OPT) benchmark to further evaluate model performance across a broader range of tasks. NL4OPT was created to explore methods for extracting the structure and formulation of optimization problems from their natural language descriptions. The primary goal of the competition is to make optimization solvers more accessible and user-friendly by enabling non-experts to interact with them using natural language \cite{tang2024orlm}.
The benchmark is divided into two sub-tasks:
(1)Entity recognition and labeling, which involves identifying and tagging the semantic entities (e.g., variables, objectives, and constraints) that represent components of the optimization problem.
(2)Meaning representation generation, where the model converts the detected entities into a structured logical form that can be transformed into an input suitable for commercial solvers\cite{tang2024orlm}.
The first task helps reduce ambiguity in the problem description by explicitly tagging relevant components. The second task focuses on building an intermediate representation of a linear programming (LP) model, which is then compiled into a solver-compatible format.
To evaluate model performance, we used GPT-4 to convert the logical forms into executable code, solve the resulting optimization problems, and verify the correctness of the solutions. These verified results were treated as ground truth. It is worth noting that approximately 15\% of the examples were discarded due to failed conversions or unsolvable formulations\cite{nl4opt}.

\subsubsection{ComplexOR}

Large Language Models (LLMs) have emerged as powerful tools for a wide range of natural language processing (NLP) tasks, including mathematical reasoning, plan generation, and code synthesis. This paper investigates the use of LLMs for automatic modeling and programming of complex operations research (OR) problems, with the aim of reducing reliance on domain experts and enabling broader application of optimization techniques across industrial sectors.
To address this objective, the authors propose Chain-of-Experts (CoE)—the first LLM-based multi-agent cooperative framework designed to enhance reasoning capabilities for complex OR tasks. Within this framework, each agent is assigned a specific role and equipped with domain-specific knowledge in OR. A central conductor agent is introduced to coordinate the interactions among agents through a combination of forward thought construction and backward reflection mechanisms.
To support the development and evaluation of CoE, the authors present a new benchmark dataset, ComplexOR, which includes a diverse set of challenging OR problems. This dataset is intended to advance research in LLM-based OR modeling and support future work within the academic and industrial communities. ComplexOR comprises 37 problems sourced from academic papers, textbooks, and real-world industry scenarios. The problem set covers a broad range of topics, including supply chain optimization, scheduling, and warehousing logistics.

The dataset was constructed with the assistance of three operations research specialists, who manually annotated each problem with precise model formulations and verified the correctness of the generated code using at least five test cases per problem. The annotation process spanned approximately one month, ensuring both depth and accuracy. Experimental results indicate that CoE significantly outperforms existing LLM-based approaches on both the LPWP and ComplexOR benchmarks, establishing a new state-of-the-art in automatic OR modeling\cite{xiao2023chain}.

\subsubsection{Mamo-EasyLP}

The Mathematical Modeling (Mamo) benchmark introduces a new approach to evaluating large language models (LLMs) in the context of mathematical optimization. The benchmark’s optimization component is divided into two segments: Easy LP, which contains 652 high school-level mixed-integer linear programming (MILP) problems, and Complex LP, which includes 211 undergraduate-level problems that integrate both linear programming (LP) and MILP.

Unlike traditional evaluation methods that focus solely on the accuracy of final solutions, Mamo adopts a process-oriented perspective. This benchmark emphasizes an in-depth assessment of the modeling strategies employed by LLMs, rather than merely verifying the correctness of their answers. By examining the steps and reasoning LLMs follow to formulate and solve optimization problems, Mamo pioneers a novel evaluation paradigm that transcends result-oriented assessment.

This shift in focus highlights the importance of understanding the inherent modeling capabilities of LLMs and provides a more nuanced and comprehensive analysis of their problem-solving behavior. The benchmark enables researchers to evaluate whether models are capable of identifying variables, constraints, and objectives from natural language descriptions—key steps in formulating optimization problems.

Mamo represents a significant advancement in the field by setting a new standard for evaluating LLMs in complex, real-world mathematical modeling tasks. It offers a framework that not only facilitates deeper insights into how models reason but also encourages the development of LLMs that are more aligned with expert-level problem formulation and interpretation. As such, Mamo points to a promising direction for future research in the intersection of artificial intelligence and operations research\cite{huang2024mamo}.

\subsubsection{COPT}

Hans Mittelmann's benchmark is used to test different solvers in solving various category of OR questions. Cardinal Optimizer(COPT) outperformed other commercial solvers like CPLEX and Gurobi in different question types. 
Cardinal Optimizer (COPT) is a high-performance mathematical programming solver designed for efficiently handling large-scale optimization problems. It has outperformed other commercial solvers, such as CPLEX and Gurobi, across various problem types. Given the nature of the benchmarks used, a powerful solver capable of addressing diverse optimization scenarios is essential. COPT's support for Linear Programming (LP), Mixed Integer Programming (MIP), Mixed Integer Linear Programming (MILP), and Mixed Integer Convex Quadratically Constrained Programming (MIQCP) makes it a strong candidate for our choice of optimizer\cite{ge2022cardinal}.

\subsection{LLMs and Optimization}

Currently, there are several approaches to utilise LLMs for Supply Chain Optimization.

OptiGuide \cite{li2023large} framework was designed for users to try out what-if analysis and bridge the gap between engineers and decision-makers. It will accept user’s input questions, generate Python codes to set up additional constraints, prompt an optimization solver to resolve the newly-formulated problem, and present the response to users. Overall, the accuracy with OptiGuide + GPT-4 is 93\%, while with GPT-4 it is only 59\%. However, Optiguide has limitations in which LLM generates totally wrong codes due to the string matching mistakes. A pinpoint of problems within LLM is needed.

Similarly, in OptiMUS \cite{ahmaditeshnizi2023optimus}, LLM was used to formulate, generate code and evaluate the processes. A new dataset, NLP4LP, which consists of challenging optimization problems (54 LP problems and 13 MILP problems) that simulate real-world problems was also purposed. This dataset was later used with NL4OPT and ComplexOR to test the effectiveness of OptiMUS. The accuracy of GPT-4 + OptiMUS is 72\% on NLP4LP, far better than other techniques like CoE \cite{xiao2024chainofexperts}. For cases of failure, the authors group the reasons into three categories:
\begin{itemize}
    \item Missing or wrong constraints because OptiMUS generates wrong constraints.
    \item Incorrect optimization method was chosen.
    \item Coding error due to incorrect codes generated. 
\end{itemize}

\subsubsection{Results from prior studies}
In contrast to the previously discussed strategies, ORLM \cite{tang2024orlm} adopts a fundamentally different approach. Rather than constructing a framework on top of existing large language models, ORLM focuses on fine-tuning LLMs to improve their performance on operations research tasks.

To support this approach, a synthetic dataset was created to cover a wide range of scenarios and difficulty levels. This dataset was used during the training phase to enable the model to better generalize across various optimization contexts. For evaluation, the authors employed the IndustryOR benchmark to assess the effectiveness of the fine-tuned LLM in solving real-world OR problems. This methodology emphasizes adapting the internal parameters of the model to enhance its optimization capabilities, rather than relying solely on external prompting or cooperative agent structures.
\begin{table}[h!]
\centering
\begin{tabular}{||c | c | c | c |c ||} 
 \hline
 \textbf{Model/Method} & \textbf{NL4OPT} & \textbf{IndustryOR} & \textbf{MAMOEasyLP} & \textbf{ComplexOR}\\ [0.5ex] 
 \hline\hline
 Llama-3.1-Instruct & 38.7\% & 13.0\% & 35.1\% & 20.8\% \\ 
 \hline
 DeepSeek-V2-Chat & 66.5\% & 16.0\% & 60.5\% & 32.7\% \\
 \hline
 Qwen2-Instruct & 72.6\% & 18.0\% & 79.9\% & 29.0\% \\
 \hline
 Mistral-Nemo & 14.6\% & 7.0\% & 19.4\% & 3.7\% \\[1ex] 
 \hline
 ORLM-Llama-3 & 85.7\% & 27.0\% & 81.4\% & 37.4\% \\ 
 \hline
 ORLM-DeepSeek-V2-Math & \textbf{86.5}\% & \textbf{33.0}\% & 82.2\% & 37.9\% \\
 \hline
 ORLM-Qwen2.5 & 86.1\% & 25\% & \textbf{85.2}\% & \textbf{44.1}\% \\
 \hline
 ORLM-Mistral & 84.4\% & 27\% & 81.4\% & 32.0\% \\[1ex]
 \hline
\end{tabular}
\caption{Initial Result Table}
\label{table:1}
\end{table}

As shown in Table \ref{table:1}, the fine-tuned LLM based on the DeepSeek-V2 model achieves the highest accuracy of 86.5\% on the NL4OPT benchmark, surpassing the performance of OptiMUS combined with GPT-4. This result highlights the potential of the DeepSeek series and motivates further exploration of other DeepSeek models for optimization-related tasks.

\lstdefinestyle{mypython}{
    language=Python,
    basicstyle=\ttfamily\small,
    keywordstyle=\color{blue},
    commentstyle=\color{gray},
    stringstyle=\color{red},
    backgroundcolor=\color{black!5},
    frame=single,
    breaklines=true,
    numbers=none
}

\section{Methodology}

To systematically evaluate the effectiveness of the DeepSeek-R1 model, a two-stage evaluation process is conducted. First, the model is assessed on the aforementioned benchmark datasets to establish its baseline performance across various operations research tasks. Following this, a comprehensive error analysis is carried out to construct a detailed error taxonomy, enabling a deeper understanding of the model’s limitations and failure modes.

Based on the insights gained from this analysis, several targeted strategies are proposed to enhance the model’s performance. These strategies aim to address specific categories of errors and improve the overall reliability and accuracy of the model in solving complex OR problems.

A detailed breakdown of the evaluation and improvement pipeline is provided below:

\begin{itemize}
    \item \textbf{Dataset Selection and Initial Model Evaluation.} The NL4OPT, IndustryOR, ComplexOR, and EasyLP datasets are used to establish the baseline accuracy of the DeepSeek-R1 model. These datasets were selected for their strong relevance to both natural language processing and mathematical optimization tasks. Evaluating DeepSeek-R1 on this diverse set of benchmarks provides an initial measure of the model’s performance \cite{tang2024orlm}. Establishing a baseline is a critical step, as it serves as a control for assessing the model’s capabilities prior to any modifications such as pretraining or fine-tuning. This enables a meaningful comparison when evaluating the impact of subsequent enhancements.
    \item \textbf{Error Analysis.}  Understanding the reasons behind model failures is essential for identifying areas of improvement. To analyze these errors systematically, an open coding strategy is adopted, as outlined by Khandkar \cite{khandkar2009open}. This method involves labeling concepts, defining categories, and developing relationships based on shared characteristics. By categorizing errors according to their similarities, a clearer picture of the model’s limitations can be formed. This structured analysis supports the development of targeted solutions and provides a framework for evaluating future iterations of the model.
    \item \textbf{Approaches to improve the model.}
    \begin{itemize}
        \item \textbf{LLM-as-a-Judge.} In this approach, DeepSeek-R1 is assigned the role of a judge to critically evaluate its own generated responses. The model assesses the correctness of the initial mathematical formulation and code. If it identifies errors or inconsistencies in the response, it attempts to diagnose the reason for the failure. Based on this self-reflection, the model then generates a revised mathematical model along with an updated code solution. This iterative process allows the LLM to refine its output through internal evaluation, enabling improved accuracy without the need for external feedback or supervision.\cite{li2024generation}. 
        \item \textbf{Tool calling for API reference.} Previous research \cite{zhang2025llmhallucinationspracticalcode} investigated the potential of using Retrieval-Augmented Generation (RAG) to retrieve relevant code snippets based on user prompts, thereby aiding the code generation process. LLMs with a set of tools, enabling them to interactively request relevant information from these tools in response to the ongoing conversation. This interaction has the potential to mitigate hallucinations during code generation, as the LLM can leverage the tool-provided information to make more informed decisions regarding API usage. The ReAct framework \cite{yao2023reactsynergizingreasoningacting} further explores optimal strategies for enabling LLMs to engage with these tools to enhance their reasoning and decision-making. In our approach, we adopt the ReAct framework to facilitate tool interaction and improve the LLM’s performance.
        \item \textbf{Multi-agent Framework.} The multi-agent framework leverages the collective intelligence of multiple large language models (LLMs) while preserving the unique capabilities of each individual agent. This design results in a sophisticated and powerful system capable of addressing complex problems that may be difficult for a single model to solve alone. In this framework, each agent is assigned a specialized role—such as mathematical modeling, code generation, or verification—and contributes to the task based on its area of expertise. Through structured collaboration and communication, the agents collectively work toward achieving the overall objective, mimicking a division-of-labor approach to problem-solving. \cite{li2024survey}
    \end{itemize}
\end{itemize}

Specifically, we will conduct ablation studies to understand the performance improvement contributed by each individual component in the above benchmarks. We will then integrate these components to identify the overall efficiency and evaluate it. 

\section{Experiments and Results}

\subsection{Evaluation Metrics}
To better simulate practical development scenarios, this study utilizes a set of coding tasks derived from real-world operations research (OR) repositories, based on established benchmarks. The IndustryOR benchmark consists of 100 real-world OR problems collected from a diverse range of industrial applications. ComplexLP contains 221 challenging linear programming (LP) problems that are considered difficult to solve, while EasyLP includes 652 relatively straightforward LP questions. Additionally, NL4OPT features 245 problems designed to evaluate the model's ability to understand and translate natural language descriptions into mathematical optimization models.

These benchmarks encompass problems across various OR domains, including transportation, aviation, simulation, and logistics. Each scenario in the datasets includes a natural language question description, a ground-truth optimal value, the variables used in the model, and their corresponding values. The accuracy of each benchmark is measured based on the number of correctly solved problems, allowing for a standardized comparison of model performance across different types and complexities of OR tasks.

\subsection{Preliminary result}
In the preliminary experiments, a total of 1,208 operations research (OR) questions were evaluated using DeepSeek-R1. Each question was paired with a prompt and a corresponding ground-truth solution. The accuracy of the model was determined by counting the number of problems it successfully solved, based on whether its output matched the expected optimal value. This initial evaluation provided a baseline measurement of the model's performance across a diverse set of OR tasks.
\begin{singlespace}
\subsubsection*{An example in the Benchmark successfully solved by DeepSeek-R1}
\begin{tcolorbox}[
    title=\textbf{An Expected Training Example},
    colframe=black!50, colback=black!5,
    breakable  
]
\textbf{Input - Natural Language Problem:} \\
The Zhang family has 6 children, Harry, Hermione, Ron, Fred, George, and Ginny. The cost of taking Harry is \$1200, Hermione is \$1650, Ron is \$750, Fred is \$800, George is \$800, and Ginny is \$1500. Which children should the couple take to minimize the total cost of taking the children? They can take a maximum of 4 children on the upcoming trip. Ginny is the youngest, so the Zhang family will definitely take her. If the couple takes Harry, they will not take Fred because Harry doesn't get along with him. If the couple takes Harry, they will not take George because Harry doesn't get along with him. If they take George, they must also take Fred. If they take George, they must also take Hermione. Although this will cost them a lot of money, the Zhang family has decided to take at least three children.

\bigskip
\textbf{Target - Mathematical Model and Program}
\textbf{Mathematical Model}
\textbf{Variables:}
\begin{itemize}
    \item $x_H, x_R, x_W,x_F, x_G, x_GI$: 0-1 variables indicating whether Harry, Hermione, Ron, Fred, George and Ginny are selected, respectively.
    \item $Z$: Total cost incurred.
\end{itemize}

\bigskip

\textbf{Objective and Constraints:}

\text{Minimize}
\[
\quad Z=1200x_H + 1650x_R + 750x_W + 800x_F + 800x_G +1500x_{GI}\\
\]

\text{Subject to:}
\begin{align*}
    \begin{cases}
        x_H + x_F  &\leq 1 \quad \text{(Mutual exclusion for Harry and Fred)} \\
        x_H + x_G  &\leq 1 \quad \text{(Mutual exclusion for Harry and George)} \\
        x_G - x_F  & < 1 \quad \text{(Fred and George must go together)} \\
        x_H - x_G  & < 1 \quad \text{(Hermione and George must go together)} \\
        x_H, x_R,x_W, x_F, x_G, x_{GI} &\in \{0,1\} \quad \text{(Binary constraints)} \\
        x_H + x_R + x_W + x_F + x_G + x_GI&\leq 4 \quad \text{(At most four children must be taken )} \\
        x_H + x_R + x_W + x_F + x_G + x_GI&\geq 3 \quad \text{(At least three children must be taken)} \\
        x_{GI} &= 1 \quad \text{(Ginny must go)}
    \end{cases}
\end{align*}

\bigskip
\textbf{Program using COPT Solver}
\begin{lstlisting}[style=mypython]
import coptpy as cp
from coptpy import COPT
# Create a COPT environment
env = cp.Envr()
# Create the model
model = env.createModel("ZhangFamilyVacation")
# Add decision variables
x_H = model.addVar(vtype=COPT.BINARY, name="x_H")  # Harry
x_R = model.addVar(vtype=COPT.BINARY, name="x_R")  # Hermione
x_W = model.addVar(vtype=COPT.BINARY, name="x_W")  # Ron
x_F = model.addVar(vtype=COPT.BINARY, name="x_F")  # Fred
x_G = model.addVar(vtype=COPT.BINARY, name="x_G")  # George
x_GI = model.addVar(vtype=COPT.BINARY, name="x_GI")  # Ginny
# Set the objective function: minimize total cost
model.setObjective(1200*x_H + 1650*x_R + 750*x_W + 800*x_F + 800*x_G + 1500*x_GI, sense=COPT.MINIMIZE)
# Add constraints
model.addConstr(x_GI == 1, name="GinnyMustBeTaken")
model.addConstr(x_H + x_F <= 1, name="HarryNotWithFred")
model.addConstr(x_H + x_G <= 1, name="HarryNotWithGeorge")
model.addConstr(x_G <= x_F, name="GeorgeImpliesFred")
model.addConstr(x_G <= x_R, name="GeorgeImpliesHermione")
model.addConstr(x_H + x_R + x_W + x_F + x_G + x_GI >= 3, name="AtLeastThreeChildren")
model.addConstr(x_H + x_R + x_W + x_F + x_G + x_GI <= 4, name="AtMostFourChildren")
# Solve the model
model.solve()
# Output results
if model.status == COPT.OPTIMAL:
    print("Minimum total cost: {:.2f} dollars".format(model.objval))
    print("Children taken:")
    if x_H.x == 1:
        print("- Harry")
    if x_R.x == 1:
        print("- Hermione")
    if x_W.x == 1:
        print("- Ron")
    if x_F.x == 1:
        print("- Fred")
    if x_G.x == 1:
        print("- George")
    if x_GI.x == 1:
        print("- Ginny")
else:
    print("No optimal solution found.")
\end{lstlisting}
\end{tcolorbox}
\end{singlespace}
Secondly, all instances of hallucinations in the generated code were systematically documented, including the specific locations where the hallucinated content appeared, as multiple hallucinations can occur within a single code sample. The identified problematic outputs were manually analyzed by an experienced Python developer, who defined and organized them into a hallucination taxonomy. Similar types of hallucinations were grouped together to form an initial classification system that reflects the nature and implications of different hallucination patterns observed in code generated by DeepSeek-R1.

After establishing the taxonomy and categorization criteria, the remaining code samples were annotated by the Python experts according to the defined categories. In cases where new types of hallucinations were discovered that were not captured by the existing taxonomy, they were reviewed and assigned to the most appropriate category. This iterative process ensured that the taxonomy remained comprehensive and adaptable. A detailed breakdown of each hallucination category, along with representative examples of the most frequently occurring hallucinations, is presented in Table \ref{table:2}.

\begin{table}[h!]
\centering
\begin{tabular}{||c | c | c ||} 
 \hline
 \textbf{Atrribute Error} & \textbf{Syntax Error} & \textbf{Logical Error} \\ [0.5ex] 
 \hline\hline
 Invalid attribute name 'update' & invalid syntax & coptcore.CoptError\\ [1ex] 
 \hline
\end{tabular}
\caption{Top Coding Hallucination Generated for each category}
\label{table:2}
\end{table}

\begin{figure}[htbp]
    \centering
    \begin{tikzpicture}
        \begin{axis}[
            ybar,
            bar width=25pt,
            width=0.95\textwidth,
            height=0.25\textheight,
            ymin=0, ymax=40,
            enlarge x limits=0.2,
            ylabel={Problemsets},
            ylabel style={font=\small},
            xtick=data,
            symbolic x coords={Attribute Error, Syntax Error, Logical Error},
            xticklabel style={rotate=0, font=\small},
            tick label style={font=\small},
            legend style={at={(0.5,-0.2)}, anchor=north, legend columns=4, font=\small},
            nodes near coords,
            every node near coord/.append style={font=\tiny}
        ]

        \addplot+[fill=blue!60] coordinates {
            (Attribute Error, 23)
            (Syntax Error, 2)
            (Logical Error, 2)
        };
        \addplot+[fill=red!60] coordinates {
            (Attribute Error, 26)
            (Syntax Error, 0)
            (Logical Error, 36)
        };
        \addplot+[fill=green!60] coordinates {
            (Attribute Error, 25)
            (Syntax Error, 2)
            (Logical Error, 5)
        };
        \addplot+[fill=orange!60] coordinates {
            (Attribute Error, 20)
            (Syntax Error, 0)
            (Logical Error, 2)
        };

        \legend{IndustryOR, EasyLP, ComplexOR, NL4OPT}
        \end{axis}
    \end{tikzpicture}
    \caption{Error type distribution across four datasets}
    \label{figure:3}
\end{figure}
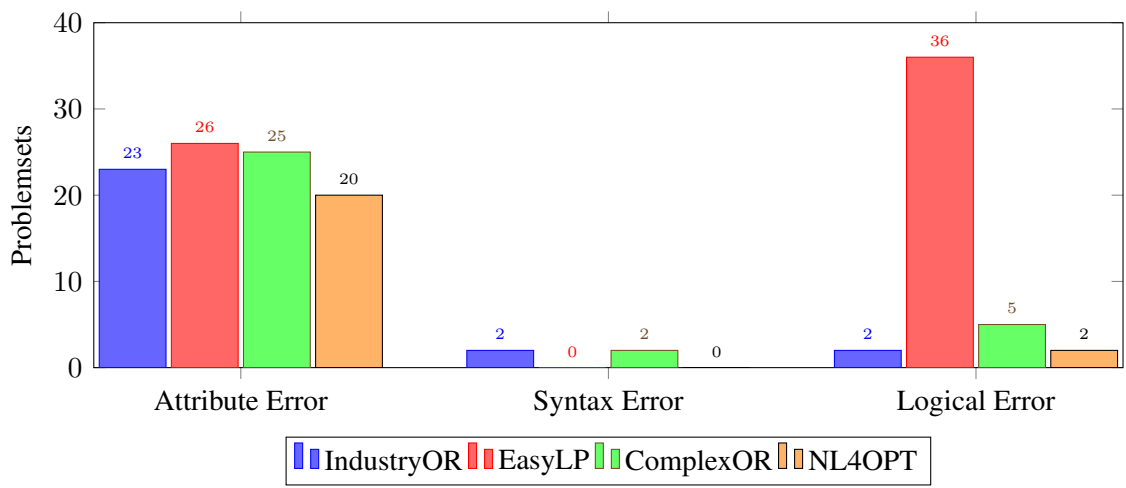

After categorizing the errors, we plotted the categories against each benchmark in Figure \ref{figure:3}.

\begin{itemize}
    \item{Attribute Error (65.7\%):} Attribute errors are one of the most common issues observed in LLM-generated code, typically resulting from incorrect or inappropriate use of application programming interfaces (APIs). These errors occur when the code attempts to access attributes or methods that do not exist within the specified object. As a result, the code fails at runtime, preventing successful execution and ultimately causing the task to remain unsolved.
    \begin{tcolorbox}[title=A sample of LLM Generation for Attribute Error, sharp corners=southwest]
    \begin{singlespacing}
    \begin{lstlisting}[style=mypython]
env = cp.Envr()
model = env.createModel("TSP")
edges = list(itertools.combinations(cities, 2))
x = {(i, j): model.addVar(vtype=COPT.BINARY, name=f"x_{i}_{j}") for (i, j) in edges}
model.update()
    \end{lstlisting}
    \textbf{Correct formulation:}
    \begin{lstlisting}[style=mypython]
env = cp.Envr()
model = env.createModel("TSP")
edges = list(itertools.combinations(cities, 2))
x = {(i, j): model.addVar(vtype=COPT.BINARY, name=f"x_{i}_{j}") for (i, j) in edges}
\end{lstlisting}
    \end{singlespacing}
\textbf{\textcolor{red}{Model class has no update method}}
\end{tcolorbox}
    \item{Syntax Error (2.80\%):} Syntax errors occur when the generated code violates Python's grammatical rules. These errors typically prevent the interpreter from executing the program, as the code cannot be parsed correctly. Such issues often arise from missing or misplaced symbols, such as parentheses, colons, or indentation. In the example below, the error is caused by an incorrect number of closing brackets, which disrupts the structure of the expression and leads to a failure during execution. 
    \begin{tcolorbox}[title=A sample of LLM Generation for Syntax Error, sharp corners=southwest]
    \begin{singlespacing}
    \begin{lstlisting}[style=mypython]
source_inflow = cp.quicksum(
variables[(i,0)] for i in edges if any(j == 0 for (j, cap) in edges[i])
model.setObjective(source_outflow - source_inflow, sense=COPT.MAXIMIZE)
\end{lstlisting}
\textbf{Correct formulation:}
    \begin{lstlisting}[style=mypython]
source_inflow = cp.quicksum(
variables[(i,0)] for i in edges if any(j == 0 for (j, cap) in edges[i]))
\end{lstlisting}
    \end{singlespacing}
\textbf{\textcolor{red}{Missing bracket}}
\end{tcolorbox}
    \item{Logical Error (31.5\%):} Logical errors occur when code executes without raising any syntax errors but produces incorrect results due to flawed reasoning or incorrect assumptions in the implementation.These errors are often more difficult to detect, as the code runs successfully but fails to achieve the desired outcome. For instance, in the following example, DeepSeek-R1 incorrectly infers the presence of an attribute - assuming that y.y exists simply because x.x does - demonstrating a misunderstanding of the object's structure.

    \begin{tcolorbox}[title=A sample of LLM Generation for Logical Error, sharp corners=southwest]
    \begin{singlespacing}
    \begin{lstlisting}[style=mypython]
if model.status == COPT.OPTIMAL:
    print(f"Optimal number of cars: {x.x}")
    print(f"Optimal number of buses: {y.y}")
    print(f"Minimum total pollution: {model.objval} units")
else:
    print("No optimal solution found")
\end{lstlisting}
\textbf{Correct formulation:}
    \begin{lstlisting}[style=mypython]
print(f"Optimal number of buses: {y.x}")
\end{lstlisting}
    \end{singlespacing}
\textbf{\textcolor{red}{y.y is not an attribute for y}}
\end{tcolorbox}
\end{itemize}

Based on the established hallucination taxonomy, a further analysis was conducted to examine the distribution of hallucination types across the four benchmarks. As illustrated in Figure \ref{figure:3}, the performance of the model varies across benchmarks, revealing distinct patterns in the types of hallucinations generated. Notably, Attribute Errors emerge as the most prevalent hallucination type across all benchmarks, indicating a consistent challenge in API usage and object referencing. Syntax Errors, on the other hand, occur relatively infrequently and are often limited to minor formatting or structural issues.

An interesting observation is that Logical Errors are more prominent in the EasyLP benchmark compared to the others. This may be attributed to the significantly larger number of problem instances within the EasyLP dataset, which increases the likelihood of exposing reasoning flaws in the model's output. These findings provide insight into how hallucination patterns are influenced by both dataset complexity and problem scale, offering valuable guidance for future model improvements and benchmark design.

\begin{figure}[h]
\centering
\begin{minipage}{0.45\textwidth}
\centering
\textbf{Overall accuracy distribution for IndustryOR: 37.0\%}
\begin{tikzpicture}[scale=0.6]
\pie[
    text=legend, 
    sum=auto, 
    color={
        red!70,    
        orange!80, 
        violet!60, 
        yellow!70, 
        green!60   
    }
]{23/Attribute Error, 2/Syntax Error, 2/Logical Error, 7/Results not Optimal, 66/Optimal Solution}
\end{tikzpicture}
\end{minipage}
\hfill
\begin{minipage}{0.45\textwidth}
\centering
\textbf{Overall accuracy distribution for NL4OPT: 78.8\%}
\begin{tikzpicture}[scale=0.6]
\pie[
    text=legend, 
    sum=auto, 
    color={
        red!70,    
        orange!80, 
        violet!60, 
        yellow!70, 
        green!60   
    }
]{26/Attribute Error, 36/Logical Error, 15/Results not Optimal, 168/Optimal Solution}
\end{tikzpicture}
\end{minipage}

\caption{Comparison of error distributions across IndustryOR and NL4OPT benchmarks.}
\end{figure}
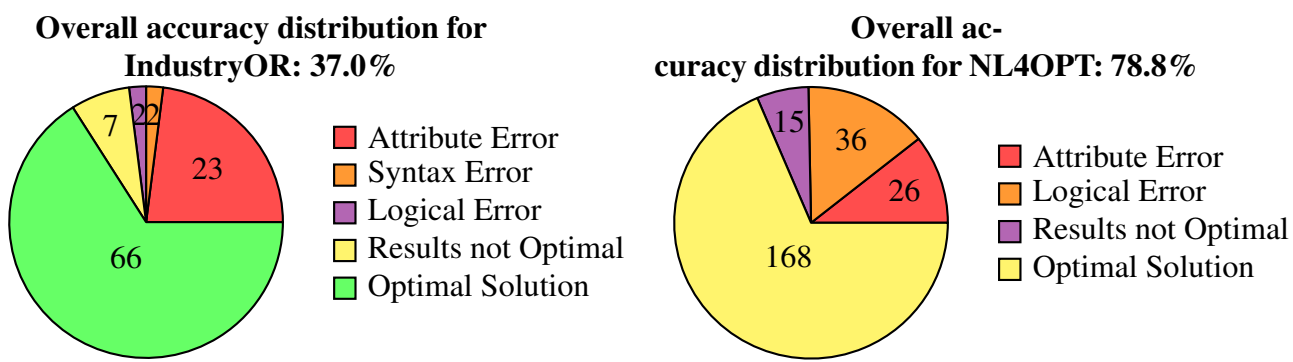

\begin{figure}[h]
\centering
    \begin{minipage}{0.45\textwidth}
    \centering
    \textbf{Overall accuracy distribution for EasyLP: 72.1\%}
    \begin{tikzpicture}[scale=0.6]
        \pie[text=legend, sum=auto,color={
        red!70,           
        orange!80,        
        violet!60,        
        yellow!70,        
        green!60          
    }]
            {25/Attribute Error, 2/Syntax Error, 36/Logical Error, 5/Results not Optimal, 584/Optimal Solution}
    \end{tikzpicture}
    \end{minipage}
    \hfill
    \begin{minipage}{0.45\textwidth}
    \centering
    \textbf{Overall accuracy distribution for ComplexOR: 53.6\%}
    \begin{tikzpicture}[scale=0.6]
        \pie[text=legend, sum=auto,color={
        red!70,           
        orange!80,        
        violet!60,        
        yellow!70,        
        green!60          
    }]
            {20/Attribute Error, 2/Logical Error, 45/Results not Optimal, 144/Optimal Solution}
    \end{tikzpicture}
    \end{minipage}

\caption{Comparison of error distributions across EasyLP and ComplexOR benchmarks.}
\end{figure}
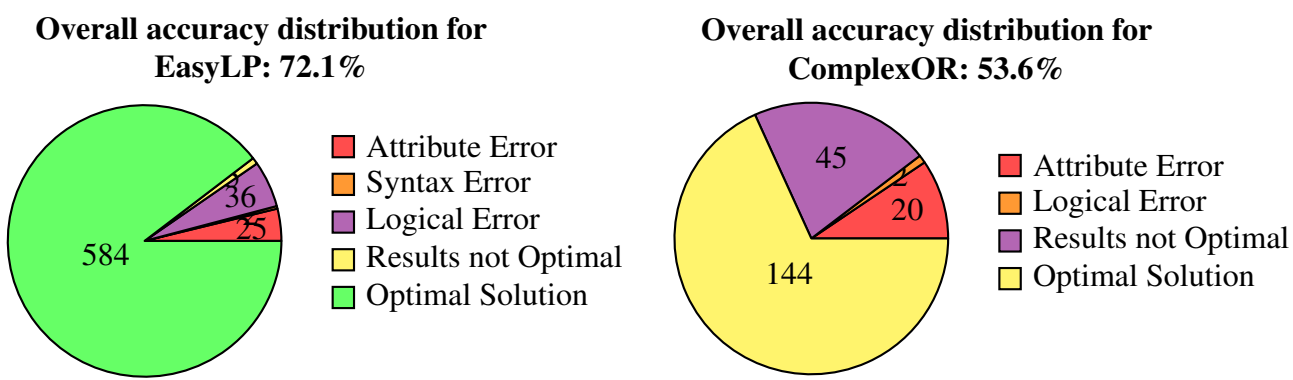

\subsection{Ablation study for Mitigation}

To evaluate the effectiveness of each proposed methodology, it is necessary to apply them individually to the DeepSeek-R1 model and assess their impact on performance. Each methodology is reviewed and analyzed in terms of its contribution to improving the model's ability to solve operations research problems.

Following the application of each approach, the correctness of the re-reasoned outputs produced by DeepSeek-R1 is evaluated using established benchmark datasets. This evaluation is conducted using the pass@k metric, a standard coding accuracy measure for large language models. As introduced by Kulal et al. \cite{kulal2019spoc}, pass@k assesses the correctness of a model’s output based on whether at least one of the top-k generated solutions passes all predefined test cases. The formula for pass@k is provided below.

\[
\text{Pass@}k = \mathbb{E} \left( 1 - \frac{\binom{n - c}{k}}{\binom{n}{k}} \right)
\]
k is the code solutions per problem, n is the number of coding solutions, and c is the number of passed solutions out of n. In this study, pass@1 is used as the primary accuracy metric, meaning that only the top-1 generated solution for each problem is considered. If this solution passes all test cases, it is counted as a successful solve.

To reduce the number of times the optimization solver runs, we evaluate each mitigation strategy on a randomly sampled 10\% subset of the dataset, repeat the process five times, and report the average pass@1 score.

Following the application of the mitigation methods, an overall improvement in model performance is observed across all four benchmarks. After obtaining the individual accuracy results for each methodology, the outcomes are further analyzed to assess the specific effects of each mitigation strategy.

Based on these insights, a practical framework is proposed to guide users in applying DeepSeek-R1 to real-world operations research (OR) problems. This framework offers a structured approach for improving problem formulation, reducing hallucinations, and increasing solution accuracy using large language models.

Table \ref{table:3} summarizes the modeling accuracy across all four benchmarks using various mitigation strategy.

\begin{table}[h!]
\centering
\begin{tabular}{||c | c | c | c | c ||} 
 \hline
 \textbf{Accuracy/Mitigation Method} & \textbf{IndustryOR} & \textbf{EasyLP} &\textbf{NL4OPT} &\textbf{ComplexOR} \\ [0.5ex] 
 \hline
 Baseline DeekSeek-R1 model & 37.0\% & 72.1\% & 78.8\% & \textbf{53.6}\% \\
 \hline
 R1+LLM-as-a-Judge & \textbf{50.0}\% & \textbf{81.3}\% & 92.3\% & 36.7\% \\
 \hline
 R1+Tool Calling & 38.0\% & 35.9\% & 46.1\% & 39.8\% \\
 \hline
 R1+LLM-as-a-Judge + FSL & 33.3\% & \textbf{81.3}\% & \textbf{95.8}\% & 26.3\% \\
 \hline
 R1+Multi-agent Framework & 0.0\% & 50.8\% & 52.0\% & 4.55\% \\
 \hline\hline
\end{tabular}
\caption{Compilation of Accuracy generated from different Mitigation Method}
\label{table:3}
\end{table}

\subsection{LLM-as-a-Judge}
In the work by Wu et al. \cite{wu2024meta}, the authors proposed a self-improvement framework in which a large language model (LLM) is trained to evaluate its own outputs and leverage this self-assessment to enhance its judgment capabilities. This unsupervised learning approach demonstrated promising results, with the LLaMA3-8B model improving its ability to judge and follow instructions from 22.9\% to 39.4\%. Inspired by these findings, this study adopts the LLM-as-a-Judge paradigm and applies it to the DeepSeek-R1 model for the task of evaluating its own reasoning and code generation outputs.

To implement this, a specially designed Judge Prompt is first fed into DeepSeek-R1, alongside its original baseline outputs. This prompt instructs the model to reflect on the correctness of its previous solution, identify any errors, and, if necessary, revise the mathematical model and code accordingly. The Judge Prompt used in this process is shown below.
\begin{singlespace}
\begin{tcolorbox}[
    title=\textbf{Judge Prompt},
    colframe=black!50, colback=black!5,
    breakable  
]
Below is an operations research question and the answer generated. Evaluate the responses and judge the correctness of the response based on the context. Evaluate the correctness of the generated mathematical model and Python code and come up with corrected version if the formulation is incorrect.
\end{tcolorbox}
\end{singlespace}

After applying the Judge Prompt to DeepSeek-R1, 10\% of data from each benchmark was sampled and tested over 10 repetitions, due to time and cost constraints. The results showed a significant improvement in accuracy for three of the benchmarks.
\begin{itemize}
\item{IndustryOR:} The accuracy improved from 37\% to 50.0\% after applying the LLM-as-a-Judge strategy. This approach successfully eliminated Logical Errors and Syntax Errors, demonstrating the model's ability to reflect on and revise its own reasoning. However, Attribute Errors still persisted, primarily due to hallucinations involving incorrect API usage. For instance, the model continued to invoke non-existent methods, such as update, without recognizing that the object in question does not support such an attribute.
While the method does not eliminate all types of errors, the results indicate that enabling the model to review and critique its own outputs significantly improves overall accuracy and robustness.
\item{EasyLP:} The accuracy increased from 72.1\% to 81.3\% after applying the LLM-as-a-Judge approach. This method successfully eliminated Attribute Errors and significantly reduced the occurrence of Logical Errors. However, some hallucinations persisted, including the specific error discussed in Section~3.2. This suggests that while the review process does not entirely prevent the model from hallucinating, it remains an effective strategy for reducing certain categories of errors and enhancing overall output reliability.
\item{NL4OPT:} The accuracy improved from 78.8\% to 92.3\% after applying the LLM-as-a-Judge method. This approach effectively eliminated Logical Errors and Syntax Errors, and also contributed to a noticeable reduction in Attribute Errors. The resulting accuracy, approaching 100\%, highlights the effectiveness of the self-evaluation mechanism in enhancing the model’s performance and reliability.
\item{ComplexOR:} The accuracy dropped from 53.6\% to 36.7\%, which can be attributed to the randomness in the selection of sample batches from the benchmark datasets. Since 10\% of the question sets were randomly sampled, it is inevitable that some batches may contain problems that the model is less capable of solving. To ensure a fair and consistent evaluation of the mitigation methods, the same batch of questions will be used in subsequent analyses. This will allow for a more reliable assessment of the DeepSeek-R1 model’s performance and enable a closer examination of the generated solutions.
\end{itemize}
\subsection{Few-shot Learning}
To enable the DeepSeek-R1 model to better address real-world operations research (OR) problems—which are often proprietary and not publicly available—few-shot learning (FSL) was incorporated into the prompting process. FSL enables the model to learn from a small number of example prompts, allowing it to infer task-specific patterns and generate accurate predictions without the need for extensive retraining. This approach is particularly valuable in domains where data is scarce or difficult to access.

By improving the model’s ability to generalize from limited examples, FSL enhances its performance on complex, domain-specific tasks. To assess the effectiveness of this approach, FSL was integrated into the baseline DeepSeek-R1 model, and its impact on performance was systematically evaluated as part of the overall mitigation strategy\cite{xu2024does}.

\begin{singlespace}

\subsubsection*{A Few-shot Learning Prompt Example}
\begin{tcolorbox}[
    title=\textbf{A Few-shot Learning Example from ComplexOR},
    colframe=black!50, colback=black!5,
    breakable  
]
\textbf{Input - Natural Language Problem:} \\
This is a integer Liner Programming problem. Imagine you're a college student aiming to balance your diet and budget. You have identified nine different food items from your local grocery store that you can include in your menu: Chicken, Rice, Apples, Steak, Lentils, Fish, Tofu, Cheese, and Bread. Each of these foods provides varying amounts of protein, carbohydrates, and calories, and each comes with its own price. Here is the detailed nutritional content and cost for each food item: Chicken: Gives you 15 grams of protein, 18 grams of carbohydrates, and 300 calories for \$4.Rice: Offers 1 gram of protein, 25 grams of carbohydrates, and 267 calories for \$2.Apples: Provide 1 gram of protein, 21 grams of carbohydrates, and 266 calories for \$5. Steak: Contains 6 grams of protein, 3 grams of carbohydrates, and 119 calories for a higher cost of \$10.Lentils: These give 3 grams of protein, 7 grams of carbohydrates, and 166 calories for just \$2. Fish: Delivers 17 grams of protein, 13 grams of carbohydrates, and 129 calories for \$8. Tofu: Offers a substantial 18 grams of protein, 27 grams of carbohydrates, and 216 calories for \$10.Cheese: Gives 12 grams of protein, 17 grams of carbohydrates, and 76 calories for \$9.Bread: Provides 2 grams of protein, a massive 30 grams of carbohydrates, and 258 calories for \$4. Your daily dietary goal is to consume at least 90 grams of protein, 105 grams of carbohydrates, and 1805 calories. Your challenge is to figure out how to meet these nutritional requirements from the food options mentioned above while spending the least amount of money. So, what is the least amount of money you need to spend to meet your daily dietary requirements? Please note that the response should be a single answer, asking for only the optimal value.
\textbf{Target - Mathematical Model and Program}
\textbf{Mathematical Model}
\textbf{Variables:}
\begin{itemize}
    \item $x_1, x_2, x_3,x_4, x_5, x_6,x_7, x_8, x_9$: integer variables indicating whether Chicken, Rice, Apples, Steak, Lentils, Fish, Tofu, Cheese, and Bread are selected, respectively.
    \item $Z$: Total cost incurred.
\end{itemize}
\textbf{Objective and Constraints:}

\text{Minimize}
\[
Z = 4x_1 +2x_2 +5x_3 +10x_4 +2x_5 +8x_6 +10x_7 +9x_8 +4x_9
\]

\text{Subject to:}
\begin{align*}
\begin{cases}
    15x_1 +1x_2 +1x_3 +6x_4 +3x_5 +17x_6 +18x_7 +12x_8 +2x_9 &\geq 90 \quad \\ \hfill 
    \text{(Protein constraint)} \\
    18x_1 +25x_2 +21x_3 +3x_4 +7x_5 +13x_6 +27x_7 +17x_8 +30x_9 &\geq 105 \quad \\ \hfill 
    \text{(Carbohydrates constraint)} \\
    300x_1 +267x_2 +266x_3 +119x_4 +166x_5 +129x_6 +216x_7 +76x_8 +258x_9 &\geq 1805 \quad \\ 
    \hfill \text{(Calorie constraint)} \\
    x_1 ,x_2 ,…,x_9 &\geq 0 \\
    x_1 ,x_2 ,…,x_9 \in \mathbb{Z} \quad \hfill \text{(Integer constraints)}
\end{cases}
\end{align*}

\textbf{This is the sample python code using COPT solver:}
\begin{lstlisting}[style=mypython]
import coptpy as cp
from coptpy import COPT
# Create environment and model
env = cp.Envr("DietProblem")
model = env.createModel("DietProblem")
# Food data
foods = ['Chicken', 'Rice', 'Apples', 'Steak', 'Lentils', 'Fish', 'Tofu', 'Cheese', 'Bread']
cost = [4, 2, 5, 10, 2, 8, 10, 9, 4]
protein = [15, 1, 1, 6, 3, 17, 18, 12, 2]
carbs = [18, 25, 21, 3, 7, 13, 27, 17, 30]
calories = [300, 267, 266, 119, 166, 129, 216, 76, 258]
# Decision variables
x = model.addVars(foods, lb=0.0, nameprefix="x",vtype=COPT.INTEGER)
# Objective: Minimize total cost
model.setObjective(sum(cost[i] * x[foods[i]] for i in range(len(foods))), COPT.MINIMIZE)
# Nutritional constraints
model.addConstr(sum(protein[i] * x[foods[i]] for i in range(len(foods))) >= 90, "protein_req")
model.addConstr(sum(carbs[i] * x[foods[i]] for i in range(len(foods))) >= 105, "carbs_req")
model.addConstr(sum(calories[i] * x[foods[i]] for i in range(len(foods))) >= 1805, "calories_req")
# Solve the model
model.solve()
# Output the result
if model.status == COPT.OPTIMAL:
    print(f"{model.objval:.2f}")
else:
    print("No optimal solution found.")
\end{lstlisting}
\end{tcolorbox}
\end{singlespace}
Applying few-shot learning (FSL) to the reviewed answers resulted in a slight reduction in accuracy. However, this decline can be attributed to specific factors observed across the three benchmarks, which are further analyzed in the following sections. 
\begin{itemize}
\item{IndustryOR:} The accuracy decreased from 50.0\% to 33.3\% after applying the few-shot learning (FSL) technique to the regenerated responses. Although logical errors were eliminated and both attribute errors and syntax errors were reduced, the overall performance declined. This reduction in accuracy can be attributed to a shift in model behavior: previously unsolvable problems were executed without errors, but failed to produce the correct optimal values, while previously solvable problems no longer yielded optimal results.
These observations suggest that, in this case, FSL did not effectively enhance the DeepSeek-R1 model's ability to identify and follow the correct steps needed to solve the problems accurately. While FSL improved execution robustness, it did not consistently guide the model toward generating optimal solutions.
\item{EasyLP:} The accuracy remained at 81.3\% after applying the few-shot learning (FSL) technique. While FSL did not eliminate the errors introduced in the responses generated by the LLM-as-a-Judge approach—particularly attribute errors resulting from hallucinations—the model continued to produce optimal solutions for the majority of the problems. This indicates that, although FSL may not fully prevent hallucinations, it does not significantly hinder the model's overall problem-solving capability. 
\item{NL4OPT:} The accuracy increased from 92.3\% to 95.8\% after applying the few-shot learning (FSL) technique. Compared to using the LLM-as-a-Judge approach alone, FSL led to a further reduction in attribute errors, contributing to the overall performance improvement. The resulting improved accuracy approaching 100\%, may be attributed to the similarity between the few-shot examples and the problem sets, which likely enabled the model to better generalize and apply relevant reasoning patterns during problem-solving.
\item{ComplexOR:} The accuracy dropped from 36.7\% to 26.3\%. The original errors observed in the LLM-as-a-Judge approach still persist when applying Few-Shot Learning (FSL). However, there is a noticeable reduction in the number of cases where execution fails entirely. Despite this improvement, many of the successfully executed responses still fail to produce the optimal value, indicating that while FSL enhances code executability, it does not consistently improve solution quality. This highlights a key limitation in the model's reasoning and optimization capabilities under few-shot prompting.
\end{itemize}
Our experimental findings are consistent with the results reported by DeepSeek \cite{guo2025deepseek}, as we also observed a decline in performance when applying few-shot learning (FSL) across the benchmarks. These observations suggest that direct prompting—without additional in-context examples—may be more effective for the DeepSeek-R1 model when solving the given problem sets.

\subsection{Multi-agent Framework}
In the proposed Multi-Agent Framework, the system is divided into two specialized agents. The first agent, referred to as the Mathematician Agent, is responsible for interpreting the question and generating the corresponding mathematical model. The second agent, the Code Agent, takes the output from the Mathematician Agent and formulates the appropriate Python code to solve the problem. The DeepSeek-R1 model is deployed for both agents, given its demonstrated capabilities in reasoning and code generation.
\begin{singlespace}
\begin{tcolorbox}[
    title=\textbf{Mathematician Agent Prompt},
    colframe=black!50, colback=black!5,
    breakable  
]
Generates a mathematical model for the given operations research question. You are an expert mathematician specializing in operations research. Given the following question, formulate an appropriate mathematical model.
\end{tcolorbox}
\end{singlespace}
\begin{singlespace}
\begin{tcolorbox}[
    title=\textbf{Coder Agent Prompt},
    colframe=black!50, colback=black!5,
    breakable  
]
Generates Python Coptpy code from the given mathematical model. You are a Python expert specializing in optimization using Coptpy. Convert the following mathematical model into Python code using Coptpy.
\end{tcolorbox}
\end{singlespace}

However, the Multi-agent Framework does not perform well on all four benchmarks. We will propose several reasons. 
\begin{itemize}
\item{General Capabilities:} While the DeepSeek-R1 model is capable of handling reasoning tasks, it may not perform as well as other LLMs that are specifically optimized for coding. As a result, it can produce code with errors that impact overall accuracy.
\item{Lack of Reasoning Context:} The Coder Agent only has access to the mathematical model output by the Mathematician Agent and does not receive any of its underlying reasoning content. This means it must independently interpret and generate coptpy code based solely on that output. If the Mathematician Agent produces an incorrect model, the Coder Agent may struggle to design correct code to find the optimal value.
\end{itemize}

\subsection{Tool Calling}
We created a tool to assist DeepSeek-R1 when solving problems. This tool extracts coptpy's function signatures from Python interface (.pyi) files and retrieves function documentation at runtime using the inspect library. When the model attempts to use a specific method or function, it can decide whether to call the tool for assistance. If the queried function exists, the tool returns the corresponding function signature and documentation. If it doesn't, the tool suggests similarly named functions to help guide the model toward a suitable alternative.
\begin{singlespace}
\begin{tcolorbox}[
    title=\textbf{Tool Calling Prompt},
    colframe=black!50, colback=black!5,
    breakable  
]
Below is an operations research question. Build a mathematical model and corresponding python code using coptpy that appropriately addresses the question. Lookup the documentation and signatures for any coptpy functions you want to use via the provided tool.
\end{tcolorbox}
\end{singlespace}
Although Tool calling is promising in reducing Attribute Errors, the performance of DeepSeek-R1 across the benchmarks is not ideal.
\begin{itemize}
\item{Hallucinations:} Although the DeepSeek-R1 model demonstrates strong performance on mathematical and programming tasks, it remains prone to hallucinations. In our experiments, we observed that the model occasionally hallucinates the output of a tool, claiming to have invoked a function or library even when the tool was never actually called. This behavior undermines the reliability of the generated solutions, particularly in code that relies on external function outputs.
\item{Confidence:} The model often exhibits high confidence in its generated responses and, as a result, rarely invokes external tools during the code generation process. Even when prompted explicitly to use the tool more frequently, this intervention did not result in a significant improvement in model performance. These findings suggest that while the model appears capable, its confidence may lead to over-reliance on internal reasoning and reduce its willingness to utilize available external resources.
\end{itemize}

\subsection{Overall accuracy}
From the above ablation studies, we identified that LLM-as-a-Judge has shown constant improvement using the benchmarks.
\begin{table}[h!]
\centering
\begin{tabular}{||c | c | c | c | c ||} 
 \hline
 \textbf{Accuracy/Mitigation Method} & \textbf{IndustryOR} & \textbf{EasyLP} &\textbf{NL4OPT} &\textbf{ComplexOR} \\ [0.5ex] 
 \hline
 Baseline DeekSeek-R1 model & 37.0\% & 72.1\% & 78.8\% & 53.6\% \\
 \hline
 ORLM-LLama-3 & 27.0\% & 81.4\% & 85.7\% & 37.4\% \\
 \hline
 DeepSeek-R1 + LLM-as-a-Judge & \textbf{50.0}\% & 81.3\% & \textbf{92.3}\% & 36.7\% \\
 \hline
 GPT4 & 28.0\% & 66.5\% & 47.3\% & 14.6\% \\[1ex]
 \hline\hline
\end{tabular}
\caption{Comparison of Accuracy on the Benchmarks by different LLM}
\label{table:4}
\end{table}
Compared to other large language models, DeepSeek-R1 combined with the LLM-as-a-Judge strategy has demonstrated promising potential. It achieved the highest accuracy on both the IndustryOR and NL4OPT benchmarks, outperforming other baseline models. Additionally, it attained comparable accuracy on the EasyLP benchmark when compared to the fine-tuned ORLM-LLaMA-3 model, highlighting its competitiveness even without task-specific fine-tuning.
However, there is still rooms for improvement. Although R1 model is capable of solving questions, it is still prone to hallucinations with lesser frequency. 
 \begin{tcolorbox}[title=A sample of LLM Generated Code for Logical Error after LLM-as-a-Judge, sharp corners=southwest]
    \begin{singlespacing}
    \begin{lstlisting}[style=mypython]
if model.status == COPT.OPTIMAL:
    print("Optimal solution found:")
    print(f"x = {x.x}, y = {y.x}, z = {z.z}")
    print(f"Minimum total hours: {model.objval}")
else:
    print("No optimal solution found.")
\end{lstlisting}
\textbf{Correct formulation:}
    \begin{lstlisting}[style=mypython]
print(f"x = {x.x}, y = {y.x}, z = {z.x}")
\end{lstlisting}
\textbf{\textcolor{red}{z.z is not an attribute for z}}
\end{singlespacing}
\end{tcolorbox}
In the example above, the DeepSeek-R1 model correctly identifies y.x as a valid attribute, demonstrating an understanding of object structure. However, it simultaneously hallucinates the existence of z.z, incorrectly assuming it to be a valid attribute as well. Nevertheless, this example suggests that the R1 model shows signs of learning from prior mistakes and possesses the capability to reduce hallucinations over time, even if such improvements are not yet consistent across all cases.

\section{Discussion}

The project began with a comprehensive literature review of the supply chain industry, with a focus on identifying real-world challenges commonly faced by companies. Based on this analysis, the scope was refined to explore the application of large language models (LLMs) in addressing operations research (OR) problems. During the review, research areas related to time series forecasting were deliberately excluded, as they did not exhibit substantial improvements in predictive accuracy within the context of our objectives.

In addition to DeepSeek-R1, the DeepSeek-V3 model was also evaluated across the selected benchmarks. Preliminary results indicated that V3 demonstrated slightly better performance than R1 in scenarios involving Tool Calling. However, despite this improvement, the overall performance of V3 remained lower than the R1 model enhanced with the LLM-as-a-Judge strategy. These findings further reinforce the effectiveness of incorporating self-evaluation mechanisms to boost model accuracy in complex OR tasks.

\section{Conclusion}
In this paper, the reasoning capabilities of the DeepSeek-R1 model were validated through its application to operations research (OR) tasks. This project represents a novel contribution to the emerging intersection between large language models (LLMs) and supply chain optimization. As this is still a relatively underexplored area, there is limited existing literature, positioning this research as one of the pioneering efforts in integrating LLMs with real-world OR problem-solving.

The study focused on conducting an ablation analysis of the base DeepSeek-R1 model by applying several mitigation strategies, including Few-Shot Learning (FSL), LLM-as-a-Judge, Tool Calling, and a Multi-Agent Framework. The effectiveness of each method was systematically evaluated, with individual accuracy improvements documented and analyzed.

Through extensive numerical experiments, it was demonstrated that DeepSeek-R1 combined with the LLM-as-a-Judge strategy significantly outperforms other LLMs, achieving an accuracy of 92.3\% on the NL4OPT benchmark, compared to 85.7\% achieved by the fine-tuned ORLM model. Furthermore, the study revealed the complementary strengths of FSL, Tool Calling, and the Multi-Agent Framework, reinforcing the base model’s capability to solve complex, real-world OR problems.

These findings validate the performance and reliability of the DeepSeek-R1 model and offer practical strategies for deploying LLMs in operational settings. Ultimately, this research marks an important step toward advancing optimization automation through LLMs, contributing an OR-driven perspective to the growing body of work on language model applications. DeepSeek-R1 thus demonstrates strong potential as a reliable and efficient solution for decision-making in real-world supply chain environments.

\subsection{Limitations}
Due to funding limitations, it was not feasible to deploy GPT-4o within the Multi-Agent Framework. As a result, GPT-4o could not be utilized to evaluate the correctness of the reasoning generated by the DeepSeek-R1 model. This constraint limited the ability to compare outputs against a more powerful external judge and restricted the exploration of hybrid frameworks that combine multiple LLMs with complementary strengths.

\subsection{Future Applications}
The findings of this research have important implications for fields that demand high accuracy and domain-specific reasoning, such as healthcare, finance, and supply chain optimization. Notably, the demonstrated performance of the DeepSeek-R1 model—achieved at a relatively low cost—offers a practical solution for companies seeking to reduce operational expenses while maintaining a high standard of output quality.

In addition, a preliminary investigation was conducted to analyze the model’s behavior during tool calling. The results revealed that DeepSeek-R1 often queries the same set of API information across different problems and does not query all the tools necessary for executing the complete set of APIs included in the generated code. This inconsistency contributes to the persistence of attribute errors in the model’s outputs.

While existing frameworks such as ReAct \cite{yao2023reactsynergizingreasoningacting} have proposed strategies to guide LLMs in making more effective use of external tools, challenges remain. Recent work by Jin et al. \cite{jin2025searchr1trainingllmsreason} explores the use of reinforcement learning to reward LLMs for leveraging search engines, offering a promising direction for future research. Similar reward-based approaches could be adapted to encourage models like DeepSeek-R1 to selectively and effectively invoke tools only when appropriate, thereby improving reliability and reducing hallucination.
\clearpage

\section{Appendix}
\subsection{Appendix A: Generating responses from DeepSeek-R1}
\singlespacing
\begin{lstlisting}[style=mypython]
TEMPLATE_q2mc_en = r"""
Below is an operations research question and answer generated. Evaluate the responses and judge the correctness of the response based on the context, Evaluate the correctness of the generated mathematical model and Python code and come up with corrected version if the formulation is incorrect.
# Question:
{Question}

# Response:
"""

def call_deepseek(prompt):
    client = OpenAI(api_key="", base_url="https://api.deepseek.com")
    # for prompt in prompt: #call each prompt 
    while True:
        try:
            response = client.chat.completions.create(
                model="deepseek-reasoner",
                messages=[
                    {"role": "system", "content": "You are a helpful assistant"},
                    {"role": "user", "content": prompt}, # ''literal string, 
                ],
                stream=False 
            )
            reasoning_content = response.choices[0].message.reasoning_content
            content = response.choices[0].message.content     
            return reasoning_content, content
        except Exception as e: #capture all exception
            print(f"OpenAI API returned an API Error: {e}")
            time.sleep(10)    #sleep for 10s

def main(dataset_name, save_dir, verbose, generated_jsonl_file, index=None):
    assert dataset_name is not None 
    assert save_dir is not None
    os.makedirs(save_dir, exist_ok=True)
    
    # Load the entire dataset
    ds = datasets.load_dataset('json',data_files=dataset_name)  # This loads all splits (e.g., 'train', 'test', etc.)
    
    # Since the dataset may contain multiple splits, you must handle that properly.
    # We'll assume it's a single split or handle it generically.
    if isinstance(ds, dict):
        # Assuming we're dealing with a single split (e.g., 'train')
        dataset_split = list(ds.values())[0]
    else:
        dataset_split = ds
    
    # Slice dataset if index is provided
    if index is not None:
        total_size = len(dataset_split)
        index = total_size #np.random.choice(total_size, size=25, replace=False)
        # start_index, end_index = index
        # dataset_split = dataset_split.select(range(start_index, end_index))  # Slice the dataset
    
    sample = []
    for example in dataset_split:
        assert "en_question" in example and "en_math_model_coptpy_code" in example and "reasoning_content" in example
    
    # Concatenate "en_question", "en_math_model_coptpy_code", and "reasoning_content" into the prompt
        prompt = TEMPLATE_q2mc_en.replace(
            "{Question}", 
            f"{example['en_question'].strip()} {example['en_math_model_coptpy_code'].strip()} "
        ).strip()
        
        # Create the example_t dictionary without the 'prompt' key
        example_t = {k: v for k, v in example.items() if k != "prompt"}
        example_t["prompt"] = prompt
        
        # Append the modified example to the sample list
        sample.append(example_t)
        
    print(f"Loaded dataset from {dataset_name}. Sample size: {len(dataset_split)}")

    # Define the output file path
    save_file = os.path.join(save_dir, generated_jsonl_file)
    prompts = [example["prompt"] for example in sample]

    # Process each prompt and example, calling Deepseek API and saving the results
    for prompt, example in zip(prompts, sample):
        reasoning_content, content = call_deepseek(prompt)
        
        # Prepare the final example for saving
        example_t = {k: v for k, v in example.items()}
        example_t["q2mc_en_prompt"] = prompt 
        example_t["judge"] = content 
        example_t['reasoning_content'] = reasoning_content
            
        if verbose:
            print("-" * 20 + " Prompt " + "-" * 20)
            print(prompt)
            print("-" * 20 + " Reasoning Content " + "-" * 20)
            print(reasoning_content)
            print("-" * 20 + " Content " + "-" * 20)
            print(content)
            print("-" * 80)
        
        dump = json.dumps(example_t, ensure_ascii=False)
        with open(save_file, 'a', encoding='utf-8') as fw:
            fw.write(dump + "\n")

    def parse_args():
    parser = argparse.ArgumentParser()
    parser.add_argument("--dataset_name", type=str, required=True)
    parser.add_argument("--save_dir", type=str, required=True)
    parser.add_argument("--generated_jsonl_file", type=str, required=True)
    parser.add_argument("--verbose", action="store_true")
    parser.add_argument("--index", type=str, default=None, help="Index range for slicing dataset (start, end) as a string")
    return parser.parse_args()
    
    main("CardinalOperations/MAMO", "complex_lp", "./data", True, "complexOR.jsonl")
\end{lstlisting}

\subsection{Appendix B: Generating Accuracy} 
\begin{lstlisting}[style=mypython]
import subprocess
import os
import json
import tempfile
import concurrent.futures
import argparse
import sys

from collections import Counter

ADD_SCRIPT = '\nif model.status == COPT.OPTIMAL:\n    print(f"Just print the best solution: {model.objval}")\nelse:\n    print("No Best Solution")'

#parse 8 means 8 values and choose the major occurance for the answer
def majority_voting(pred_answers):
    # Count occurrences of each item in the list
    count = Counter(pred_answers)
    # Find the answer with the maximum count
    max_count = max(count.values())
    # Extract all answers with the maximum count
    possible_answers = [answer for answer, cnt in count.items() if cnt == max_count]
    # Return the first answer with the maximum count
    return possible_answers[0]


def compile_script(script_content, timeout=300):
    # Ensure the target directory exists
    target_dir = "./eval_execute"
    os.makedirs(target_dir, exist_ok=True)

    # create a temporary file to store script
    with tempfile.NamedTemporaryFile(
        delete=False, suffix=".py", dir=target_dir
    ) as tmp_file:
        tmp_file_name = tmp_file.name
        tmp_file.write(script_content.encode())

    try:
        # Running Python on the temporary script file with a time limit
        process = subprocess.run( #call programe will start a process, use another python file to run
            [sys.executable, tmp_file_name], #temperary file generated to run in the pipe
            text=True,
            stdout=subprocess.PIPE, #output is the new input
            stderr=subprocess.PIPE,
            timeout=timeout,
            check=True,
        )

        # If compilation is successful, return the output and a success message
        execution_result = process.stdout 
        # find the start of output we want to extract
        # == -1 when we cannot find "Just print the best solution:" in the code
        # != -1 means we find "Just print the best solution:" in the output
        execution_best_solution_start_pos = execution_result.find(
            "Just print the best solution:"
        )

        print(execution_result)

        # if we find such an output
        if execution_best_solution_start_pos != -1:
            execution_best_solution = (
                execution_result[execution_best_solution_start_pos:]
                .replace("Just print the best solution:", "")
                .strip()
            )
            execution_best_solution_end_pos = execution_best_solution.find("\n")
            if execution_best_solution_end_pos != -1:
                execution_best_solution = execution_best_solution[
                    :execution_best_solution_end_pos
                ]
            execution_state = "Execution Successful and Best Solution Found"
        else:
            # when no such output in the console
            if "No Best Solution" in execution_result:
                execution_best_solution = "No Best Solution"
                execution_state = "Execution Successful but No Best Solution Found"
            # when something unexpected happened
            else:
                execution_best_solution = None
                execution_state = "Execution Suceessful but Out of Expectation"
    # when timeout expired
    except subprocess.TimeoutExpired as e:
        # If compilation time exceeds the limit, kill the process and return a failure message
        execution_result = e.stdout
        execution_best_solution = None
        execution_state = "Execution Failed: Timeout"
    # when failed to call
    except subprocess.CalledProcessError as e: #nvr return 0(success), return a value
        # If compilation fails for other reasons, return the error output
        execution_result = e.stdout  #output stream of e
        execution_best_solution = None
        execution_message= e.stdout + '\n' + e.stderr
        execution_state = f"Execution Failed: {execution_message}" 
    #look at error message
    # this block will be executed after all the above
    finally:
        # Clean up the temporary file
        os.remove(tmp_file_name)

    execution_output = {
        "execution_result": execution_result,
        "execution_best_solution": execution_best_solution,
        "execution_state": execution_state,
    }
    return execution_output #return dictionary


def main(args):
    # Check version
    # - ensure dependencies to run code is installed
    process = subprocess.run( #check coptpy is installed
        ["python", "--version"],
        text=True,
        stdout=subprocess.PIPE,
        stderr=subprocess.PIPE,
        timeout=10,
        check=True,
    )
    print(process.stdout)
    process = subprocess.run(
        ["python", "-c", "import coptpy"],
        text=True,
        stdout=subprocess.PIPE,
        stderr=subprocess.PIPE,
        timeout=10,
        check=True,
    )
    print(process.stdout)
    print("check coptpy installed.")

    # Load scripts to compile
    early_failed = [] #failed to extract code from generated answer
    to_run = [] #can run but maybe fail
    with open(args.input_file) as fd:
        # load each generated output
        for line in fd:
            example = json.loads(line)

            # extract code field
            code_field = None #extract from the generated code
            for key in example.keys():
                if "coptpy_code" in key:
                    code_field = key
                    break
            assert code_field is not None

            output = example[code_field]
            example_t = {k: v for k, v in example.items()}

            # failed to extract code, append this to early fail
            start = output.find("```python")
            if start == -1:
                execution_output = {
                    "execution_result": "Execution Failed: No code",
                    "execution_best_solution": None,
                    "execution_state": "Execution Failed: No code",
                }
                example_t.update(execution_output)
                early_failed.append(example_t)
                continue

            # extract script
            end = output.find("```", start + 9)
            script = output[start:end].replace("```python", "")

            # no code is generated
            if script.strip() == "":
                execution_output = {
                    "execution_result": "Execution Failed: No code",
                    "execution_best_solution": None,
                    "execution_state": "Execution Failed: No code",
                }
                example_t.update(execution_output)
                early_failed.append(example_t)
                continue

            script += ADD_SCRIPT #check if the coptpy output is the optimal solution 
            # print(script)
            # raise
            example_t["to_run_script"] = script #dictionary, script is the value for to_run_script, add everyline into the to_run_script and run the script later
            to_run.append(example_t)
    # to_run = to_run[0:10]
    print(f"len(to_run): {len(to_run)}")
    # raise

    # Function to process each example
    def process_example(example):
        execution_output = compile_script(example["to_run_script"]) 
        # print(f"-" * 20 + "compilation result" + "-" * 20)
        # print(execution_output["execution_result"])
        # print("-" * 10)
        # print(execution_output["execution_best_solution"])
        # print("-" * 10)
        # print(execution_output["execution_state"])
        example.update(execution_output)
        return json.dumps(example, ensure_ascii=False) #change to string

    # create a thread pool that executes submitted tasks; a lot of threads that capable of doing task
    with concurrent.futures.ThreadPoolExecutor( 
        max_workers=args.max_workers #how many workers
    ) as executor:
        # Submitting all the tasks to the executor
        # - this allows tasks to run concurrently
        future_to_example = {
            executor.submit(process_example, example): example for example in to_run
        } #give a function to submit to process

        # Writing the results to file as they are completed
        with open(args.output_file, "w", encoding="utf-8") as fw:
            for example in early_failed:
                dump = json.dumps(example, ensure_ascii=False)
                fw.write(dump + "\n")

            # iterate over all the completed tasks
            for future in concurrent.futures.as_completed(future_to_example):
                try:
                    result = future.result()
                    fw.write(result + "\n")
                except Exception as exc:
                    print(f"An error occurred: {exc}") #if the future throw exception, go error
                    continue

    print("Execution completed.")

    # if question_field ("en_question") and answer_field ("en_answer") are given
    #calculated accuracy matric
    if (args.question_field is not None) and (args.answer_field is not None):
        question2pred_answers = {}
        question2gt_answers = {}  # the "label"
        judges = [] 

        # load answer from jsonl file
        with open(args.output_file, "r") as fd:
            for line in fd:
                example = json.loads(line)
                question = example[args.question_field]
                if question not in question2pred_answers:
                    question2pred_answers[question] = []
                if question not in question2gt_answers: #groud truth
                    question2gt_answers[question] = []

                gt_answer = example[args.answer_field]
                question2gt_answers[question].append(gt_answer)

                pred_answer = example["execution_best_solution"]
                question2pred_answers[question].append(pred_answer)

        k = -1
        for question, pred_answers in question2pred_answers.items():
            k = len(pred_answers)

            gt_answers = question2gt_answers[question]
            assert len(set(gt_answers)) == 1 #ensure only 1 gt answer
            gt_answer = gt_answers[0]

            # iterate over all predicted answers (there is just one prediction in our case)
            is_anyone_match = False
            for pred_answer in pred_answers:
                # if both of them are "No Best Solution"
                if gt_answer == "No Best Solution":
                    if pred_answer is not None and pred_answer == gt_answer:
                        is_anyone_match = True
                        break
                else:
                    # check if two answers are close enough, if so, we treat it as a match
                    gt_answer = round(float(gt_answer))
                    if pred_answer is not None and pred_answer != "No Best Solution":
                        pred_answer = round(float(pred_answer))
                        if gt_answer == 0:
                            close_enough = (
                                abs(pred_answer) <= args.numerical_err_tolerance 
                            )
                        else:
                            close_enough = (
                                abs((pred_answer - gt_answer) / gt_answer)
                                <= args.numerical_err_tolerance
                            )
                        if close_enough:
                            is_anyone_match = True
                            break

            if is_anyone_match:
                judges.append(1) #match judge=1, not match judge=0
            else:
                judges.append(0)

            if args.verbose:
                print("-" * 60)
                print("-" * 20 + "question" + "-" * 20)
                print(question)
                print("-" * 20 + "pred_answers" + "-" * 20)
                print(pred_answers)
                print("-" * 20 + "gt_answer" + "-" * 20)
                print(gt_answer)
                print("-" * 20 + "judge" + "-" * 20)
                print(is_anyone_match)

        # calculate accuracy
        acc = sum(judges) / len(judges)
        metrics = {f"pass@{k}": acc}

        # majority voting -> it uses the answer with the largest count as the final answer
        # - not useful in our scenario where there is only one answer
        if args.majority_voting:
            mj_judges = []
            for question, pred_answers in question2pred_answers.items():
                k = len(pred_answers)

                gt_answers = question2gt_answers[question]
                assert len(set(gt_answers)) == 1
                gt_answer = gt_answers[0]

                pred_answers_t = []
                for pred_answer in pred_answers:
                    if pred_answer is None:
                        continue
                    try:
                        pred_answer = round(float(pred_answer))
                        pred_answers_t.append(pred_answer)
                    except:
                        pred_answers_t.append(pred_answer)
                if pred_answers_t != []:
                    mj_answer = majority_voting(pred_answers_t)
                else:
                    mj_answer = None

                # similar logic as above
                is_mj_match = False
                if gt_answer == "No Best Solution":
                    if mj_answer is not None and mj_answer == gt_answer:
                        is_mj_match = True
                else:
                    gt_answer = round(float(gt_answer))
                    if mj_answer is not None and mj_answer != "No Best Solution":
                        if gt_answer == 0:
                            close_enough = (
                                abs(mj_answer) <= args.numerical_err_tolerance
                            )
                        else:
                            close_enough = (
                                abs((mj_answer - gt_answer) / gt_answer)
                                <= args.numerical_err_tolerance
                            )
                        if close_enough:
                            is_mj_match = True
                if args.verbose:
                    print(
                        f"gt_answer: {gt_answer}; pred_answers_t: {pred_answers_t}; mj_answer: {mj_answer}; is_mj_match: {is_mj_match}"
                    )

                if is_mj_match:
                    mj_judges.append(1)
                else:
                    mj_judges.append(0)

            # calculate a majority acc
            mj_acc = sum(mj_judges) / len(mj_judges)
            metrics[f"mj@{k}"] = mj_acc

        if args.output_file.endswith(".json"):
            metrics_file = args.output_file.replace(".json", ".metrics.json")
        elif args.output_file.endswith(".jsonl"):
            metrics_file = args.output_file.replace(".jsonl", ".metrics.json")
        else:
            metrics_file = args.output_file + ".metrics.json"
        with open(metrics_file, "w") as fw:
            dump = json.dumps(metrics, indent=4)
            fw.write(dump)
            print(dump)


def parse_args():
    parser = argparse.ArgumentParser()
    parser.add_argument("--input_file", type=str)
    parser.add_argument("--output_file", type=str, default=None)
    parser.add_argument("--timeout", type=int, default=600)
    parser.add_argument("--max_workers", type=int, default=16)
    parser.add_argument("--verbose", action="store_true", default=False)
    parser.add_argument("--majority_voting", action="store_true", default=False)
    parser.add_argument("--question_field", type=str, default=None)
    parser.add_argument("--answer_field", type=str, default=None)
    parser.add_argument("--numerical_err_tolerance", type=float, default=0.05)
    return parser.parse_args()


if __name__ == "__main__":
    args = parse_args()
    main(args)
\end{lstlisting}

\subsection{Appendix C: Evaluating Errors} 
\begin{lstlisting}[style=mypython]
import pandas as pd
import json

# jsonObj = pd.read_json(path_or_buf='generated_IndustryOR.jsonl', lines=True) #add reletive path
# print(jsonObj) 

input_file = "multi-complexOR_executed.jsonl"

# Define a function to filter specific sections
def extract_sections(input_file):
    with open(input_file, "r", encoding="utf-8") as infile: #, open(output_file, "r", encoding="utf-8") as outfile:
        for line in infile:
            data = json.loads(line)  # Load each JSON object
            if data.get('execution_best_solution') is None or data.get('execution_state') != "Execution Successful and Best Solution Found":  # Filter condition, look at the value corresponding to key
                #outfile.write(json.dumps(data) + "\n")  # Write to new JSONL file
                print(data.get('execution_state'))
                print(data.get('en_math_model_coptpy_code'))
                print(data.get('prompt'))
                print("-" * 200 + "\n\n"+ "-" * 200)
extract_sections(input_file) #give key value
\end{lstlisting}

\newpage

\printbibliography

\label{appendix:tokenization}

\end{document}